\documentclass[acmlarge]{acmart}
\makeatletter
\newcommand{\myconfshort}{\acmConference@shortname}
\newcommand{\myconffull}{\acmConference@name}
\newcommand{\myconfdate}{\acmConference@date}
\newcommand{\myconfloc}{\acmConference@venue}
\AtBeginDocument{
  \fancypagestyle{firstpagestyle}{
    \fancyhead{}%
    \fancyfoot[C]{}%
  }
  \fancyhf{}
  \fancyhead[LO]{\@headfootfont\shorttitle}%
  \fancyhead[RE]{\@headfootfont\@shortauthors}%
  \fancyhead[LE]{\@headfootfont\footnotesize \myconfshort, \myconfdate, \myconfloc}%
  \fancyhead[RO]{\@headfootfont\footnotesize \myconfshort, \myconfdate, \myconfloc}%
  \fancyfoot[C]{}%
}
\makeatother
\acmBooktitle{\conffull\@ (\confshort), \confdate, \confloc}
\AtBeginDocument{%
  }

\copyrightyear{2026}
\acmYear{2026}
\setcopyright{cc}
\setcctype{by}
\acmConference[FAccT '26]{The 2026 ACM Conference on Fairness, Accountability, and Transparency}{June 25--28, 2026}{Montreal, QC, Canada}
\acmBooktitle{The 2026 ACM Conference on Fairness, Accountability, and Transparency (FAccT '26), June 25--28, 2026, Montreal, QC, Canada}
\acmDOI{}
\acmISBN{}
\usepackage{enumitem}

\usepackage{tcolorbox}
\usepackage{listings}
\tcbuselibrary{breakable, skins, listings}

\newtcblisting{promptbox}[1]{
  title={\textbf{#1}},
  breakable,
  colback=gray!8,
  colframe=gray!40,
  fonttitle=\small\bfseries,
  coltitle=black,
  boxrule=0.4pt,
  arc=3pt,
  boxsep=2pt,
  top=1pt,
  before=\par\vspace*{4pt},
  listing only,
  listing options={
    basicstyle=\ttfamily\small,
    breaklines=true,
    breakatwhitespace=true,
    columns=fullflexible,
    breakindent=2em,
  }
}

\begin{document}

\title{Who Defines "Best"? Towards Interactive, User-Defined Evaluation of LLM Leaderboards}


\author{Minji Jung}
\affiliation{%
  \institution{Yonsei University}
  \city{Seoul}
  \state{}
  \country{South Korea}}
\email{minji2744@yonsei.ac.kr}

\author{Minjae Lee}
\affiliation{%
  \institution{Yonsei University}
  \city{Seoul}
  \state{}
  \country{South Korea}}
\email{minjaelee@yonsei.ac.kr}

\author{Yejin Kim}
\affiliation{%
  \institution{Yonsei University}
  \city{Seoul}
  \state{}
  \country{South Korea}}
\email{gina261@yonsei.ac.kr}

\author{Sarang Choi}
\affiliation{%
  \institution{Yonsei University}
  \city{Seoul}
  \state{}
  \country{South Korea}}
\email{sarangchoi@yonsei.ac.kr}

\author{Minsuk Kahng}
\authornote{Corresponding author}
\affiliation{%
  \institution{Yonsei University}
  \city{Seoul}
  \state{}
  \country{South Korea}}
\email{minsuk@yonsei.ac.kr}

% \renewcommand{\shortauthors}{}


\begin{abstract}
LLM leaderboards are widely used to compare models and guide deployment decisions. However, leaderboard rankings are shaped by evaluation priorities set by benchmark designers, rather than by the diverse goals and constraints of actual users and organizations. A single aggregate score often obscures how models behave across different prompt types and compositions. In this work, we conduct an in-depth analysis of the dataset used in the LMArena (formerly Chatbot Arena) benchmark and investigate this evaluation challenge by designing an interactive visualization interface as a design probe. Our analysis reveals that the dataset is heavily skewed toward certain topics, that model rankings vary across prompt slices, and that preference-based judgments are used in ways that blur their intended scope. Building on this analysis, we introduce a visualization interface that allows users to define their own evaluation priorities by selecting and weighting prompt slices and to explore how rankings change accordingly. A qualitative study suggests that this interactive approach improves transparency and supports more context-specific model evaluation, pointing toward alternative ways to design and use LLM leaderboards.

\end{abstract}




% \received{14 January 2026}
% \received[revised]{25 March 2026}
% \received[accepted]{15 April 2026}

\maketitle

\section{Introduction}

Large Language Models (LLMs) are increasingly evaluated, compared, and selected through benchmark datasets and leaderboards. These leaderboards play a central role in shaping research narratives, deployment decisions, and public perceptions of model quality. Yet, despite their influence, most evaluation frameworks are designed and fixed by a small group of benchmark creators, while their results are consumed by a much broader and more diverse audience.

This asymmetry raises a fundamental concern: evaluation outcomes are largely determined by data composition and aggregation choices made by others, and then inherited by users whose goals, contexts, and use cases may differ substantially. Leaderboard rankings implicitly reflect which prompts are included, how frequently they appear, and how results are aggregated into a single score. Global aggregate rankings obscure substantial variation across topics, domains, or task types. For many real-world scenarios, performance on particular data subsets may matter far more than overall averages, yet such distinctions are rarely visible or actionable.

One common response to this limitation is to disaggregate evaluation results~\cite{barocas2021designing, madaio2022assessing}, for example by reporting performance across predefined slices or by using automated methods to identify meaningful subgroups in the data. While valuable, these approaches typically assume that the structure and relative importance of slices are fixed in advance. In practice, however, users often need to explore how different parts of the data contribute to overall results, and to decide for themselves which subsets deserve greater emphasis based on their context.

In this paper, we argue that LLM evaluation should support this interactive exploration and reweighting process. Rather than asking users to take a fixed global ranking, or a fixed set of disaggregated views, we propose treating leaderboard evaluation as a sensemaking activity, where users can examine data slices, adjust their relative importance, and observe how rankings change as a result.

To ground this argument, we conduct an in-depth analysis of the dataset used in the popular LMArena (formerly Chatbot Arena) benchmark~\cite{chiang2024chatbot}. It reveals substantial skews in prompt coverage and pronounced variability in model rankings across different data slices. These findings demonstrate how a single aggregate ranking can reflect the dominant composition of the dataset, while concealing trade-offs that emerge when attention shifts to specific subsets.

Building on this analysis, we design an interactive visualization tool as a design probe to explore how users might engage with disaggregated leaderboard results. The interface exposes the composition of the evaluation dataset and allows users to define important prompt slices and adjust their relative emphasis, making it possible to observe how model rankings change as users shift their attention across different subsets of data. The visualization serves as a concrete instantiation of our analysis to examine how user interaction can support more contextual interpretation of leaderboard outcomes.

We conducted a small qualitative study (N=10) to understand how practitioners would use this design in their model selection process. Participants used our system to inspect different prompt types and explore how emphasizing different slices affected rankings under their needs and constraints. These observations suggest that interactive, slice-based views can help users move beyond global rankings and interpret leaderboard results more critically.

This work makes three primary contributions:
\begin{itemize}[topsep=4pt]
\item A reframing of LLM leaderboard evaluation as an interactive process of exploring and reweighting data slices, rather than relying on fixed aggregate evaluations.

\item An empirical analysis of the LMArena leaderboard data, demonstrating how dataset composition can affect model rankings.

\item An interactive visualization interface as a design probe, with qualitative findings illustrating how more contextual and transparent interpretation of leaderboard results could work in practice.
\end{itemize}

\section{Related Work}

\subsection{Benchmarking Practices in LLM Evaluation}

While benchmarking has become the de facto standard for AI evaluation, its dominance has introduced significant practical limitations. Current benchmarks cannot represent the full scope of real-world use; they act as finite samples rather than measures of general intelligence~\cite{raji2021ai}. The overemphasis on leaderboard competition has transformed evaluation 
into what some call ``AI as a sport,'' where beating benchmark scores becomes the goal~\cite{orr2024ai}. Research efforts tend to cycle around a small number of datasets that are repeatedly reused and recycled, encouraging optimization for specific benchmarks at the expense of broader applicability~\cite{koch2021reduced}. These practices assume that benchmark test sets are representative of the real world, but when datasets reflect biased or 
incomplete representations, the evaluations themselves become systematically 
biased~\cite{buolamwini2018gender}.

To assess diverse capabilities of language models, the field has increasingly adopted large-scale, multi-task benchmarks~\cite{hendrycks2020measuring, srivastava2023beyond, chung2024scaling, suzgun2023challenging}. In parallel, evaluation practices are shifting toward holistic approaches that aim to capture a broader range of metrics and model behaviors~\cite{shevlane2023model, liang2022holistic}. For example, HELM~\cite{liang2022holistic} assesses model performance beyond accuracy, incorporating measures such as calibration, robustness, fairness, toxicity, and efficiency. However, as benchmarks expand to encompass more tasks and dimensions, their aggregate rankings become increasingly unstable and sensitive to seemingly minor changes~\cite{zhang2024inherent}.

\subsection{Preference-based Evaluation and Leaderboards}

The evaluation of LLMs has shifted from reference-based metrics toward preference-based judgments, reflecting the open-ended and subjective nature of language generation tasks. The LMSYS group popularized this approach by using pairwise preference comparisons, initially employing strong LLMs as judges to provide a scalable proxy for model quality~\cite{vicuna2023, zheng2023judging, li2024crowdsourced, dubois2024length}.
Building on this work, the same group introduced Chatbot Arena~\cite{chiang2024chatbot}, later rebranded as LMArena,\footnote{Note that the platform was again rebranded from LMArena to Arena in late January 2026. In this paper, we use LMArena to reflect the name when most of this research was conducted.} where users compare two model responses side-by-side and vote for their preferred response. These crowdsourced judgments are aggregated using an Elo-based rating system to produce a continuously updated leaderboard intended to reflect real-world human utility.

Recent studies have revealed fundamental limitations in LMArena-style platforms and preference-based evaluations.
Aggregation mechanisms such as Elo obscure performance heterogeneity across different tasks~\cite{rofin2023vote, lanctot2023evaluating, boubdir2024elo, singh2025leaderboard}, and evaluations occur disproportionately for queries that are very easy or highly objective, meaning ties are driven more by intrinsic query properties than by model capability~\cite{tang2025drawing}. The systems are also vulnerable to adversarial manipulation~\cite{min2025improving, huang2025exploring, frickprompt}.
Perhaps most critically, the preference-based evaluation tends to prioritize perceived helpfulness, while weakly constraining other important dimensions, like factual correctness, honesty, and safety~\cite{feuer2024style, chen2024humans, wu2025style}.
These limitations stem not from any specific evaluation method, but from the structural reliance on aggregated preference signals, motivating our focus on disaggregated and transparent evaluation approaches.

\subsection{Subgroup Analysis and Interactive Slicing}

Global aggregate metrics often obscure performance heterogeneity across diverse domains, demographic groups, and user intents.
Failing to disaggregate performance data erases minority group experiences and reinforces systemic social inequities~\cite{buolamwini2018gender}, which can be particularly consequential in high-stakes domains like healthcare~\cite{obermeyer2019dissecting, seyyed2021underdiagnosis}.
Aggregate metrics are structurally incapable of ensuring fairness and responsibility in AI systems, creating a critical need for disaggregated evaluation~\cite{herlihy2024structured, diaz2024scaling, khodak2024suremap, pfohl2025understanding}. 
Behavioral testing methodologies similarly emphasize 
the importance of systematic, fine-grained probing to expose localized failure 
modes~\cite{ribeiro2020beyond, wu2019errudite}.
These approaches can ensure that the benefits of AI are distributed equitably across a range of stakeholders and use cases~\cite{barocas2021designing, madaio2022assessing}.

A key insight from the human-computer interaction (HCI) community is that disaggregated evaluation must 
be interactive.
Rather than presenting users with a fixed set of slices, interactive tools allow users to define, 
explore, and compare slices that matter for their specific context~\cite{cabrera2023zeno, 
sivaraman2025divisi, kahng2024llm}. 
These systems leverage visual analytics 
techniques that help users make sense of complex patterns through iterative, 
human-guided exploration, emphasizing user agency~\cite{cabrera2019fairvis, zhang2022sliceteller, wang2024visual}.
This interactive philosophy extends beyond model evaluation to documentation.
The Interactive Model Cards suggest that static documentation is insufficient for diverse stakeholders, advocating for tools that allow users to interactively probe model boundaries~\cite{crisan2022interactive}. Our work builds on this interactive slicing approach to 
leaderboard
evaluation, a setting that has received little attention in prior work. While existing 
tools focus on analyzing individual models or pairwise comparisons, we enable users to interactively define evaluation priorities that simultaneously analyze the relative standing of many models.

\subsection{Interactive Ranking and User-Defined Evaluation}

While most leaderboards present fixed rankings that apply uniformly to all 
users, recent work has begun exploring more flexible approaches. The Prompt-to-Leaderboard framework optimizes for personalized, prompt-conditioned leaderboards~\cite{frickprompt}, and
Arena-Hard-Auto prioritizes more informative or challenging prompts through automated selection~\cite{li2024crowdsourced}.
However, these approaches still rely on predetermined notions of what makes a good evaluation. As Jury Learning demonstrates in the context of annotation~\cite{gordon2022jury}, disagreement and diverse perspectives are not noise to be eliminated but 
signal to be preserved~\cite{aroyo2015truth}.
Similarly, the composition of the prompt dataset shapes what counts as "best," and this is inherently a value-laden process.

To address the need for context-specific evaluation, we draw on principles from multi-attribute ranking visualizations~\cite{gratzl2013lineup, seo2005rank, pajer2016weightlifter, wall2017podium}. Systems like LineUp~\cite{gratzl2013lineup} demonstrate the power of visualization for interactively exploring trade-offs across multiple dimensions. 
Similarly, Dynaboard~\cite{ma2021dynaboard} enables users to customize a scoring function by adjusting weights across metrics such as accuracy, robustness, and efficiency.
While recent work has explored user-defined criteria for LLM evaluation~\cite{kim2024evallm, shankar2024validates}, we take a complementary approach. We focus on leveraging existing evaluation data, rather than asking users to define abstract criteria from scratch. 
We aim to support the fundamental visualization principle of revealing patterns through interaction, so that users can discover insights about model behavior~\cite{north2006toward}. 
This transforms leaderboard 
evaluation from passive consumption into active sensemaking, supporting more transparent and context-specific model selection.

\section{Dataset Analysis}
\label{sec:dataset-analysis}

\subsection{Dataset Properties and Topic Distribution}

\subsubsection{Dataset Overview}
We analyze the Human Preference 140K dataset,\footnote{\url{https://huggingface.co/datasets/lmarena-ai/arena-human-preference-140k}} the latest release from the LMArena platform. The dataset comprises preference-based judgments for 53 LLMs, collected over about three months from April to July 2025. On the LMArena website, users see responses from two different LLMs side-by-side and vote for their 
preference: Model A wins, Model B wins, Tie, or Both Bad. 
The dataset contains 135,634 judgments: Model A won 35.8\%, Model B won 36.7\%, and Ties (including Both Bad) accounted for 27.5\% of cases.
It also includes metadata for each prompt, such as category tags 
(e.g., whether it contains code, or is about 
mathematics) and language.

\begin{figure*}[!t]
    \centering
    \includegraphics[width=1.0\linewidth]{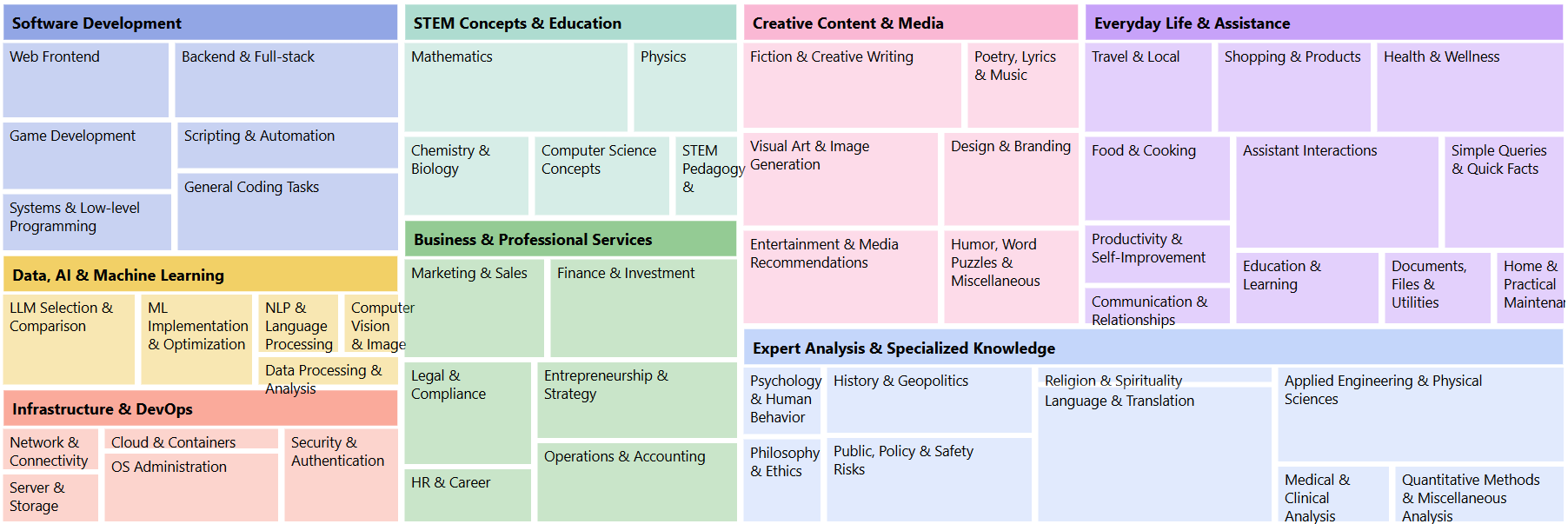}
    \caption{Treemap visualization of the topic distribution of the LMArena dataset. We construct a three-level topic hierarchy and visualize the top two levels in this figure; rectangle areas are proportional to the number of prompts in each category. The distribution reveals a clear skew toward developer- and AI-related topics, which together account for 30\% of the dataset, reflecting the interests of a specific population.}
    \label{fig:hierarchy-treemap}
\end{figure*}

\subsubsection{Semantic Topic Hierarchy}
\label{sec:categorization}

To analyze the semantic composition of the dataset, we construct a three-level topic hierarchy through a two-stage pipeline: low-level grouping via clustering, followed by higher-level organization via LLMs with manual refinement.

\textbf{Low-level clustering.} To avoid grouping prompts by lexical similarity rather than semantic intent,
we first generate a short English topic description using GPT-5 mini (prompt in Appendix \ref{box:topic hiearchy - topic generation}), following Lam et al.~\cite{lam2024concept} and Tamkin et al.~\cite{tamkin2024clio}. We then compute embeddings using OpenAI's \texttt{text-embedding-3-small} model and apply $k$-means clustering.
The choice of $k$ involves a trade-off: as $k$
increases, clusters become more descriptively specific: clusters at $k\geq400$ contain significantly more unique terms than clusters at smaller $k$ values, but each cluster's smaller sample size widens confidence intervals and increases the probability that any two models' score distributions overlap. We evaluated $k \in \{100, 200, ..., 600\}$ and found that this probability grows nearly linearly. We selected $k$ to be 400  as an empirical middle ground that preserves descriptive specificity while maintaining reasonable statistical power.

\textbf{Higher-level categories.} We organize the 400 clusters into top- and mid-level categories following LLM-based topic clustering methods where an LLM proposes candidate groupings, similar to Pham et al.~\cite{pham2024topicgpt} and Wang et al.~\cite{wang2023goal}.
We use the more capable GPT-5.2 for this more abstract grouping task (see Appendix~\ref{box:topic hiearachy - constructing higher level} for the prompt).
Since fully automated hierarchy construction~\cite{tamkin2024clio} produced categories that were not consistently coherent at the same level of abstraction, we refined the hierarchy through manual review.
Specifically, 10\% of higher-level categories are manually added, and 13\% of clusters are reassigned.
To assess robustness, we repeated the pipeline twice for the six different $k$'s and compared which prompts fell into high- vs. low-divergence mid-level categories (top and bottom 20\%); prompt-level agreement was 86\% (Cohen's $\kappa$=0.71), suggesting that divergence patterns are reasonably consistent across clustering configurations.

The final hierarchy consists of 8 top-level, 53 mid-level, and 400 fine-grained categories. Figure~\ref{fig:hierarchy-treemap} presents a treemap visualization for the distribution of the top- and mid-level categories, with rectangle areas proportional to the number of prompts in each category.
We observe three notable patterns:
\begin{enumerate}[topsep=2pt]
\item First, programming and software development 
prompts dominate the dataset, accounting for about 30\% of all prompts across multiple categories.
This reflects overrepresentation of a specific population (e.g., software developers and AI practitioners) whose needs and preferences may differ substantially from the broader user 
base. This pattern is consistent with prior analyses of LMArena and similar corpora~\cite{zhao2024wildchat, tamkin2024clio}.

\item Second, the dataset contains over 1,000 prompts that consist only of simple greetings (e.g., ``hi there'') that appear to be system tests~\cite{arena2024hard}. 
Interestingly, despite minimal differences between model outputs in many cases, users selected a winner in 79\% of these cases, suggesting arbitrary choices that add noise to aggregate rankings.

\item Third, we observe many highly repetitive prompts that sometimes form a distinct cluster in our hierarchy.
For example, ``How many `r's are there in strawberry?'' appears 205 times with only minor wording variations. This question has been used within the user community to test model behavior and is submitted by many users, which can distort rankings.
\end{enumerate}

\begin{figure*}[!t]
    \centering
    \includegraphics[width=1.0\linewidth]{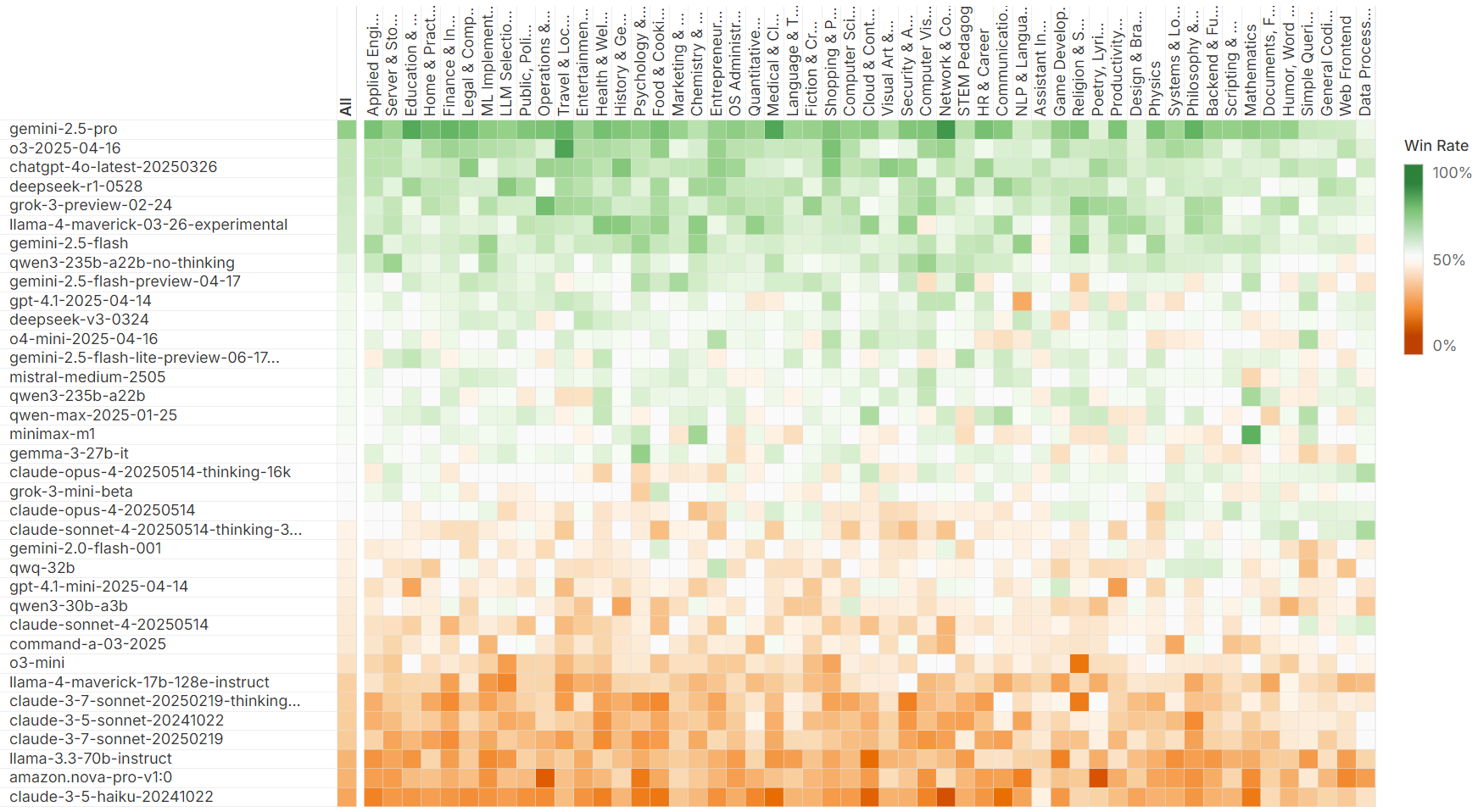}
    \caption{Heatmap visualization of model performance across mid-level prompt categories. Rows represent models with at least 4,000 evaluations, sorted by overall win rate. The first column shows each model’s overall win rate, and the remaining columns correspond to mid-level categories ordered by their Spearman correlation with the overall ranking, from left (high correlation) to right (low correlation), with Data Processing \& Analysis appearing at the far right. Notably, this lowest-correlation category shows that Claude-family models achieve relatively higher win rates compared to their overall performance. Each cell encodes the model’s win rate for the corresponding category; deviations from surrounding colors highlight category-specific performance differences, such as the high win rate of minimax-m1 in Math, visible as a dark green cell toward the middle right.}
    \label{fig:heatmap}
\end{figure*}

\subsection{Category-Specific Ranking Differences}

Model performance can vary substantially across different prompt categories. To understand this variability, we analyze how rankings change across the mid-level categories in our topic hierarchy.

\subsubsection{Which Categories Show Different Rankings?}
\label{sec:different-rankings-categories}

For each mid-level category, we compute per-model win rates and compare their ranking with the overall ranking using Spearman's rank correlation. 
We use Bayesian smoothing based on the Beta-Binomial model to account for uncertainty with small sample sizes~\cite{murphy2012machine}, and exclude ties. The top three categories with highest correlations are Applied Engineering, Home Maintenance, and Server \& Storage (over 0.93); the bottom three are \textit{Data Processing \& Analysis} (0.60), \textit{Web Frontend} (0.70) and \textit{Simple Queries \& Quick Facts} (0.76). 

To better understand these divergences, we examine \textit{Data Processing \& Analysis}, the lowest-correlation 
category, in detail. 
This category shows dramatic ranking differences.
Claude has eight models in the dataset, none ranking in the overall top 20, but in this category, three Claude models jump to the top 5.
To understand what drives these wins, we prompt an LLM 
to generate rationales for why users preferred winning responses. 
(detailed prompt is in Appendix~\ref{box:winning-model-analysis}). The LLM consistently attributes Claude's wins to correctness and precision in handling structured data, but prompts requiring 
such complexity are rare in a dataset dominated by lighter queries.

\subsubsection{Which Models Perform Differently in Specific Categories?}

The above analysis identified categories where rankings differ from the overall ranking. We now examine a finer-grained question: which specific models show unusually strong or weak performance in particular categories, even when the category as a whole may not deviate substantially?

We first visualize win rates across models and categories as a heatmap (shown in Figure~\ref{fig:heatmap}), where rows represent models and columns represent mid-level categories sorted by the Spearman's correlation values, with each cell showing the model’s win rate. While many models perform consistently well across categories, some show isolated spikes.
We quantify these deviations using a \textit{two-proportion z-test}~\cite{fleiss2013statistical}, 
comparing each model's win rate within a category to its win rate across all other categories.
Higher absolute z-scores indicate larger differences. 

The strongest deviation is observed for \texttt{minimax-m1} in the math category, where the $z$-score exceeds 8. The model achieves 138 wins and only 30 losses. At the fine-grained subcategory level, it excels particularly in algebra question solving (95\% win rate, $z = 5.43$) and advanced number theory (92\%, z = 4.16), 
which indicates that the model performs well on difficult computational tasks, rather than on simple routine problems. 

These findings reveal that models can excel on specific task types, such as complex coding tasks that emphasize correctness and precision. However, because such tasks constitute a small portion of the dataset, strong performance on them is underrepresented in aggregate rankings. This raises 
a fundamental question: can preference-based evaluation meaningfully aggregate across tasks with fundamentally different characteristics—those with deterministic answers, those involving subjective preferences, and those requiring value-dependent judgments?

\subsection{Limits of Preference-Based Evaluation Across Deterministic and Pluralistic Tasks}

To better understand when aggregate preference-based evaluation can misrepresent model behavior, we examine settings where such aggregation is most likely to break down. Rather than aiming for exhaustive coverage of all prompt types, we focus on two representative extremes: (1) prompts with deterministic, objectively correct answers; and (2) prompts involving inherently pluralistic, value-dependent judgments.

\subsubsection{Objective Tasks with Deterministic Answers}
\label{sec:deterministic-math-analysis}

We analyze 8,035 prompts from the mathematics category of the LMArena dataset, using the category metadata provided in the dataset. However, not all prompts in this category have deterministic answers.
We exclude prompts with ambiguous meaning or no deterministic answer using GPT-5.2 by asking it to mark mathematical questions with a well-defined verifiable answer (detailed prompt is in Appendix~\ref{box:math-filtering-deterministic}). We classify a total of 2,773 prompts as mathematical problems with a definitive answer.

\paragraph{Answer Correctness.}
We utilize GPT-5.2 and Gemini 3 Flash to classify model responses based on correctness by asking LLMs to assess whether each response gives a correct mathematical answer to the prompt (prompt in Appendix~\ref{box:math-correctness}). We acknowledge that LLM-based judgments can be incorrect. To mitigate this limitation, we employ two independent models, and analyze 
only
the 2,143 cases (4,286 responses) where both models agreed on correctness.

In 23\% of the cases, one model response is correct and the other is incorrect. While correctness correlates relatively highly with human preference (Spearman's rank correlation coefficient 0.71), humans selected the correct answer as the winner only 74\% of the time. Our inspection of data reveals that users made incorrect selections on problems involving complex formulas or precision-sensitive calculations, such as discount rates, taxes, or square roots. Also, we notice that in 67\% of the cases, both models' responses are correct for the same prompt. Even when both models answered correctly, humans selected a winner 56\% of the time. This indicates that getting the correct answer alone does not lead to winning, and other factors influence user judgments. We hypothesize that explanation styles might affect the preference evaluation which we investigate below.

\begin{table*}[!t]
    \centering
    \begin{tabular}{p{3.8cm} p{10.1cm} r}
    \toprule
    Name & Description & Count\\
    \midrule
    Conciseness & Delivers the core answer or solution directly with minimal explanation & 2,256 \\
    Elaboration & Expands beyond minimal answers with additional context or explanations & 2,259 \\
    Structure Richness & Organizes the response using clear formal structure, e.g., steps or tables & 2,788 \\
    Reasoning with Derivation & Shows step-by-step reasoning, intermediate steps, and derivations & 3,022 \\
    Rigorous Assumption Handling & Carefully examines assumptions, constraints, edge cases and potential ambiguities & 3,069 \\
    User-oriented Interaction & Actively engages the user with follow-up questions, clarification, etc. & 898 \\
    \bottomrule
    \end{tabular}
    \caption{Explanation style characteristics in model responses related to the mathematics category. Six characteristics were identified through analysis using GPT-5.2.}
    \label{tab:six-styles}
\end{table*}

\paragraph{Explanation Styles.}

To understand how explanation style influences preference,
we identify style characteristics with GPT 5.2 (see Appendix~\ref{box:math-explanation-style-identify} for the prompt). We extract and refine recurring traits by crafting a prompt with 100 randomly sampled examples, repeating for ten iterations, and merging the results. 
This process yields six distinct characteristics as listed in Table~\ref{tab:six-styles}.
We then use GPT-5.2 to compare model response pairs and tag each response for the presence of these characteristics (see Appendix~\ref{box:math-label-explanation} for the prompt).
For response pairs where both models provide correct answers, we measure how much these sets overlap between the two responses using Jaccard similarity.
High similarity indicates that the two responses share largely the same stylistic reasons for being preferred, whereas low similarity suggests that different stylistic factors may have influenced human judgments.

When humans selected a winner despite both model responses being correct, the average Jaccard similarity between response pairs was
0.3, which is lower than that of randomly paired responses (0.5). 
This pattern suggests that human preferences in these cases might arise not because both responses succeed in similar ways, but because one response stands out along different stylistic dimensions.
Among winning responses, the most frequent characteristics are conciseness (50\%), elaboration (48\%), and structure richness (39\%).
These results indicate that, in math-related problems, human preferences are shaped by specific combinations of explanation styles rather than correctness alone.
Consistent with this interpretation, the \texttt{minimax-m1} model, which balances concise answers with clear structure and derivational reasoning, rises from 19th globally to 1st in math, while also maintaining a high accuracy rank of 4th.

\begin{table*}[!t]
    \centering
    \begin{tabular}{lcccc}
    \toprule
    Prompt type & Total Count & Non-pluralistic & Refusal & Ratio of Non-pluralistic\\
    \midrule
    Geopolitical conflicts & 162 & 18 & 2 & 11.1\% \\
    Value judgments & 88 & 19 & 1 & 21.5\% \\
    Political or historical issues in China & 249 & 41 & 17 & 16.4\% \\
    Human rights issues & 42 & 3 & 0 & 16.4\% \\
    Political or historical evaluation & 444 & 29 & 1 & 6.5\% \\
    Future predictions & 173 & 10 & 0 & 5.7\% \\
    \bottomrule
    \end{tabular}
    \caption{Category classification of political dispute related prompts, illustrating variation across prompt types in the proportion of model responses that take a specific position (non-pluralistic) rather than presenting pluralistic viewpoints.}
    \label{tab:politics-category}
\end{table*}

\subsubsection{Value-Dependent Tasks with Pluralistic Judgments}
\label{sec:pluralistic-politics-analysis}

Unlike tasks with a single correct answer, some prompts involve inherently value-dependent judgments, where reasonable disagreement arises from differences in cultural, political, or moral perspectives.
One approach that has been explored is to train LLMs to produce pluralistic responses that present multiple viewpoints and avoid taking a specific stance~\cite{Pryzant2019AutomaticallyNS, Lahoti2023ImprovingDO, Hoffmann2025ImprovingNP}. 
However, responses in such cases may not be appropriately evaluated against a single objective standard.
To understand how such pluralistic and position-taking responses are evaluated in practice, we examine how users judge model responses to value-dependent prompts.

To empirically examine how users evaluate value-dependent prompts, we analyze 2,895 data points from the \textit{History \& Geopolitics} category from our hierarchy (Section~\ref{sec:categorization}). This analysis involves a two-stage labeling process using three LLMs (GPT-5-mini, Gemini 3 Flash, and Claude Opus 4.5),
applying majority voting
to prevent any single model's biases from being over-reflected in the process: first, we identify whether a prompt involves political or historical disputes where reasonable differences in opinion exist based on cultural and political contexts; second, for prompts identified as such, we label each model response based on whether it adopts a politically neutral stance with pluralistic viewpoints or advances a specific position (prompt in Appendix ~\ref{box:political-pluralistic-label}).

Following the above process, we identify a total of 1,158 prompts (2,316 responses) containing political or historical disputes. At the response level, 139 responses (6\%) are non-pluralistic and 22 responses (1\%) refused to answer. We measure inter-rater reliability on 278 responses (all 139 non-pluralistic responses and an equal random sample of pluralistic ones). 
Three humans (two co-authors uninvolved in this analysis and one involved) independently labeled each response, and we aggregated them via majority voting, mirroring the LLM setup.
The Krippendorff's alpha~\cite{krippendorff2018content} between the humans and LLMs is 0.726, which is reasonably high and higher than agreement among the three LLMs alone (0.656) and among the three human labelers alone (0.576),
though interpretation should account for the value-laden nature of the task.
We then define six categories of prompt types as listed in Table~\ref{tab:politics-category} by examining a 20\% sample (231 prompts) and classify those using GPT-5.2 (prompt in Appendix ~\ref{box:politics-category-classify}). The inter-rater reliability between human and LLMs is 0.850.

Among the prompt categories, two categories stand out for their high proportions of non-pluralistic model responses. First, \textit{value judgments} prompts, such as questions asking ``Which country is right: Israel or Palestine?'' elicit position-taking responses in 21.5\% of the time. Next, prompts discussing \textit{Chinese political or historical issues} (e.g. the Tiananmen Square protests, the 2022 COVID-19 protests in China, and Taiwan-China sovereignty problem) produce non-pluralistic responses in 16.4\% of the cases.

Evaluation in the presence of non-pluralistic responses becomes inherently subjective and contingent on evaluators’ values. Cases where one of the responses is non-pluralistic, excluding ties (81 prompts), pluralistic answers won 46\% of the time while non-pluralistic answers won 54\% (95\% CI: [0.44, 0.65]; binomial test against 50\%, p=0.25).
This indicates no statistically significant preference for either type, despite an expectation that pluralistic responses would be favored.
We further examine whether this instability is also reflected at the level of model rankings. The model with the largest drop in political or historical dispute related prompts compared to the overall ranking is \texttt{kimi-k2-0711-preview}, which falls by 16 places in the ranking. Although its sample size is small, it shows political bias in approximately 30\% of its preference judgments, recording the highest rate. The top three models with the highest rates of non-pluralistic responses are \texttt{kimi-k2-0711-preview}, \texttt{deepseek-r1-0528}, and \texttt{qwen3-30b-a3b}.

These findings highlight a structural limitation of LMArena’s evaluation paradigm. Because rankings are derived from aggregated human preference judgments, they collapse diverse and often conflicting value positions into a single score. As a result, models may be systematically favored when their responses align with the views of dominant evaluator groups, particularly in politically or culturally contested domains. This dynamic raises concerns that shifts in evaluator composition may directly translate into model rankings, independent of broader notions of quality or responsibility. Moreover, if models adapt strategically to such preference distributions, such as selectively adopting pluralistic framing or refusal behaviors only for sensitive topics, it may introduce additional risks, including inconsistent treatment of similar issues and the marginalization of minority perspectives. Therefore, these observations call into question the adequacy of aggregate preference-based leaderboards for evaluating models in value-dependent contexts.

\section{Interactive Visualization Interface}

Building on our analysis, we introduce an interactive visualization interface as a design probe to explore how users might 
actively shape evaluation in ways that reflect their own goals and contexts.

\subsection{Design Goals}
Our goal is not to address every limitation revealed in our analysis, but to explore a core idea through an interface that points toward a different way of engaging with leaderboard evaluation. The design is guided by three key goals:
\begin{enumerate}
    \item First, we aim to shift control over evaluation composition from benchmark designers, such as crowd users in LMArena, to stakeholders who develop LLM-based applications. Instead of relying on a fixed overall metric determined by the benchmark's original data composition, users should be able to construct their own evaluation dataset by selecting and weighting prompt categories according to their priorities.
    \item Second, we recognize that users cannot fully specify an ideal evaluation setup upfront. Therefore, the interface supports incremental sensemaking, allowing users to iteratively explore, refine, and adjust their evaluation choices through interaction.
    \item Third, rather than limiting evaluation to abstract categories and aggregate model scores, the interface emphasizes access to concrete prompt examples and to the intersection patterns between categories and models. This allows users to better understand why models perform differently across slices, not just how they rank.
\end{enumerate}

\begin{figure*}[!t]
    \centering
    \includegraphics[width=1.0\linewidth]{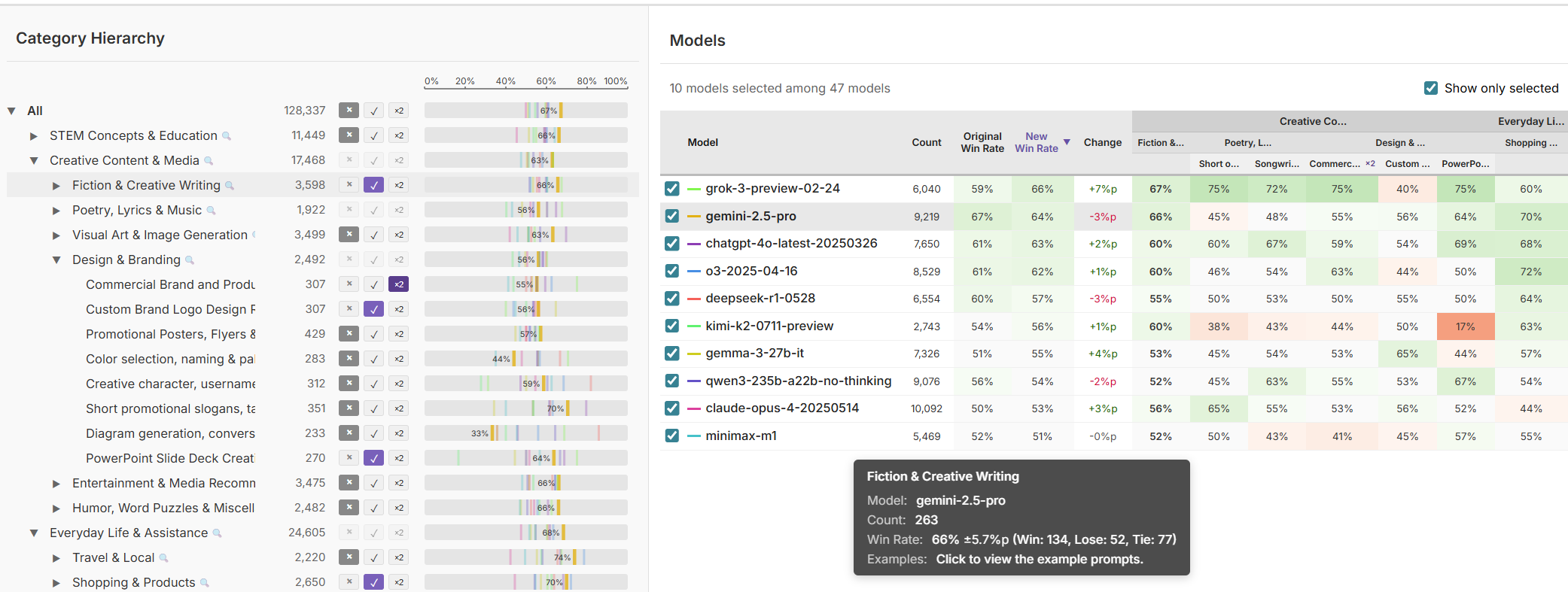}
    \caption{Interactive visualization interface for user-defined leaderboard evaluation. The interface consists of two coordinated panels: the left panel allows users to select and weight prompt categories from a hierarchical topic structure, while the right panel shows the resulting model rankings under the selected slices.
    To support sensemaking about model-category relationships, the left panel summarizes category-specific rankings using strip plots, and the right panel encodes category-level win rates as colored cells, making unusually strong or weak model-category combinations easy to identify. By clicking on individual cells, users can inspect additional details via tooltips and prompt examples (not shown).}
    \label{fig:interface}
\end{figure*}

\subsection{Interface Design}

The interface is organized into two coordinated views arranged side by side as shown in Figure~\ref{fig:interface}. The left panel presents a category selection view, while the right panel presents a model ranking view. After selecting categories in the left panel, the right panel updates to show model rankings computed over the selected slices.
The interface is designed following principles from information visualization, with an emphasis on tight coordination between views. Users can inspect model rankings while adjusting category selections, and conversely, examine how a model’s performance varies across categories while focusing on the ranking view. Rather than treating categories and models as independent dimensions, the design emphasizes revealing their intersections, helping users identify where and why ranking differences emerge. We describe each component in more detail below.

\subsubsection{Category or Slice Selection}

Prompt categories are displayed as a hierarchical tree, reflecting the topic hierarchy introduced in Section 3. Users can expand or collapse branches of the tree to select top-level or mid-level categories, and selectively exclude subcategories that are not relevant to their use case.

Categories can be incrementally added or removed, and their weights can be adjusted to reflect relative importance. Because category names alone may not sufficiently convey their content, we allow users to inspect individual prompt examples by clicking a magnifying-glass icon. This design choice is critical: by directly examining example prompts, users can assess whether a category aligns with their actual needs before including it in the evaluation.

\subsubsection{Model Ranking View}

The right panel displays the model ranking. Models are shown as rows in a table-like layout familiar from leaderboard designs. Instead of displaying only a single weighted aggregate score, the columns show performance across the selected slices, enabling users to compare how models behave across categories.

Performance values are encoded using a heatmap, making deviations from surrounding patterns visually salient. Hovering over a cell reveals tooltips that include additional details such as error ranges. Clicking on a cell reveals the underlying prompt examples associated with that model-category combination, allowing users to directly inspect cases where a model performs unusually well or poorly and to understand the reasons behind these differences.

\subsubsection{Coordinated Interactions}

A key aspect of the interface is the coordination between the views. When users focus on a particular model and examine combinations of slices, a strip-plot visualization alongside the category tree highlights how that model ranks across other categories, making relative performance differences easier to analyze.
In addition, users can revise category compositions while inspecting examples or identifying outlier patterns. Categories can be expanded, reduced, or reweighted, and these changes are reflected immediately in the ranking view. This dynamic feedback loop supports efficient iterative refinement and encourages users to treat evaluation as an exploratory, interactive process.

\subsection{Scenarios}

Consider Priya, who works on a team building an online tutoring application for teenage students. Her team plans to deploy an internally fine-tuned LLM to support personalized learning across subjects such as language, history, and science. Because using an off-the-shelf external LLM is costly, the team has developed and tuned around ten internal model variants tailored to their tasks. They now need to evaluate these variants and turn to our interface.

Priya opens the interface and begins by defining relevant categories. She excludes categories related to software development and AI that are irrelevant to their use case and instead focuses on education and subject knowledge related categories. Upon inspecting some fine-grained subcategories, she removes advanced or overly specialized topics that go beyond the level of teenage learners, while retaining those aligned with the teenage curriculum. At the same time, she assigns higher weight to categories related to broadly important capabilities such as audience-tailored explanations.

After configuring these slices, Priya examines the resulting model rankings and finds that Model \#4 (MiniMax-based) and Model \#6 (DeepSeek-based) perform best overall. A closer inspection, however, reveals a trade-off. Model \#4 performs strongly on quantitative subjects such as math and physics, but less so on history and language. Model \#6, in contrast, is somewhat weaker on some math and physics categories, but performs more consistently, especially on language tasks. Priya shares these findings with her team to focus their refinement on these two models.

This process allows Priya to move beyond relying on a generic leaderboard or overall accuracy scores. Instead, the interface supports deliberate and transparent decision-making grounded in tasks that matter to their organization, along with an understanding of the trade-offs and stability of the chosen model.

\subsection{Implementations}

The system is implemented as a web application. The frontend is built using React, with custom visualizations rendered using SVG. To avoid loading the full dataset on the client, the frontend stores only category assignments and win-loss statistics for each example. Detailed data are fetched on demand from a Python Flask backend when users request them. This architecture supports efficient scaling while enabling interactive computation in the interface.

\section{Qualitative User Study}
We conducted a qualitative user study to explore how users would engage with our proposed interactive visualization interface.

\subsection{Participants and Procedure}
We recruited ten participants via a mailing list. Participants were required to have prior experience developing or evaluating AI models for their professional work, or comparable academic experience. These criteria ensured that participants had substantial experience with LLMs and sufficient background to critically assess model behaviors. Participants came from a range of backgrounds, including AI engineers working on retrieval-augmented generation pipelines and LLM safety, a front-end engineer, an operations planning professional at a manufacturing company, a data analyst in credit assessment, a security consultant, a university lecturer, and AI/NLP graduate students. 
Participants’ years of experience averaged 5.5 years (SD: 2.9), and each participant received a gift card (KRW 50,000, approximately USD 35) for their time. 

Each session was conducted as a 45-minute, one-on-one interview via Zoom.
The session consisted of a brief pre-interview, followed by a short tutorial on the interface, a period in which participants used the system themselves, and a post-interview. 
The main task was to use the interface to identify, among ten provided LLMs, the one they considered most appropriate for a given target scenario.
Participants worked with a scenario that was either drawn from their own professional context, identified during the pre-interview, or provided by the researchers (e.g., customer service chatbot for an e-commerce website, AI-powered tutoring chatbot). Participants used the interface to complete the task while following a think-aloud protocol, and were asked to justify their final model choice. We recorded audio and screen sharing with participants' consent, transcribed all sessions, and analyzed the data using thematic analysis. The study was approved by our institution's IRB.

\subsection{Findings}

We identify five key themes that describe how participants used the interface to support model selection in their own context.

\subsubsection{Challenging and Confirming Prior Model Perceptions}
Participants encountered results that either contradicted or confirmed their prior assumptions about model performance. Several had strong preconceptions, often based on personal experience or popular discourse, about which models would excel. In some cases, they found patterns that contradict 
their expectations.
P1, who has been thinking \texttt{Claude} \texttt{Opus} as the best model based on their personal experience, discovered that \texttt{Gemini} \texttt{Pro} outperformed it on their selected categories. In addition, P2 observed that smaller or less prominent models sometimes outperformed flagship models on specific tasks, challenging the assumption that general popularity correlates with domain-specific quality. In other cases, the rankings validated prior perceptions. P7 had encountered online discussions describing \texttt{Gemini} as generally strong and \texttt{Grok} as particularly effective for creative content generation, and found these patterns reflected in the results.

\subsubsection{Refining Contextual Relevance Through Fine-Grained Category Selection.}
Participants used the hierarchical category structure to tailor evaluations to their own contexts, but often discovered that broad higher-level categories contained subcategories misaligned with their actual needs. P4 said: \textit{``At the broad category level, things seemed relevant to our task, but when I drilled into subcategories, some weren't relevant at all. Being able to see fine-grained categories helped me focus on what we actually need.''} P5, evaluating models for math education, deliberately selected subcategories targeting high school and early undergraduate levels while excluding others, as they have thought that advanced topics fell outside their intended user population. The ability to exclude irrelevant slices served as a critical aspect of the interaction.

\subsubsection{Grounding Decisions Through Example Inspection.}
Beyond adjusting category weights, participants relied on concrete prompt examples to interpret rankings and verify that categories matched their expectations. P3 said: \textit{``After selecting categories based on the types of questions, it was great that [...], and looked at examples to see specifically why that model performed better. This helped me select a model with more confidence, as I could understand why it had a higher win rate.''} P9 similarly inspected examples to understand why a model was preferred and to compare that judgment with their own assessment: \textit{``That made me think that the labelers' evaluation results were similar to my own.''} This behavior suggests that participants sought to ground quantitative metric numbers by observing behavior qualitatively through examples.

\subsubsection{Calibrating How Category-Level Results Should Count}
Participants calibrated how much different category-level results should count in their decisions by considering both statistical reliability and task importance. P2 encountered a case where \texttt{Kimi} achieved 100\% win rate in a math education category but dismissed it because the sample size was small. P10 similarly attended to sample size, deliberately adding related categories so that the recomputed win rate would be based on sufficient evidence. Beyond sample size, participants also adjusted category importance in both implicit and explicit ways. P5, despite observing that \texttt{Minimax} performed poorly on a ``comparing decimal point numbers'' subcategory, deprioritized this because \textit{``students can probably handle that on their own, so I weighted it lower in my head.''}
Conversely, P8 and P9 used the interface's double-weighting feature to make these priorities explicit, rather than relying only on mental adjustment. These behaviors reveal that participants did not passively consume the interface's outputs but actively negotiated with the data, applying domain knowledge to temper statistical signals.

\subsubsection{Reframing Leaderboards as Decision Aids.}
Interactive exploration helped participants reconsider how leaderboards fit into their decision-making process. Rather than treating rankings as definitive answers, participants began to view them as tools that support, but do not replace, their own judgment. P2 described how the interface could structure team discussions: \textit{``I would first organize the capabilities needed from the model, set priorities across categories, and then present the resulting rankings to explain why this is the best model for us.''} P4 had been skeptical of existing leaderboards, noting that models often overfit to test data. However, the ability to select fine-grained categories made the evaluation feel more relevant: \textit{``[...] being able to focus on specific categories feels much better. It resonates more with what I actually need.''} P6 distinguished between research contexts, where established leaderboards remain relevant, and deployment contexts, where task-specific evaluation becomes essential.

\section{Discussion}
\label{sec:discussion}

We discuss the implications of our findings for evaluation practice and for the design of interactive tools that support more responsible use of leaderboards.

\subsection{Leaderboards as a Sensemaking Process}

Our dataset analysis shows that model rankings in LMArena vary substantially across prompt categories. In particular, we observed sharp ranking divergences in categories such as data processing and mathematics, as well as instability in politically contested prompts. These results indicate that leaderboard rankings are not stable reflections of an underlying, unified notion of model quality, but are instead highly contingent on which subsets of data are emphasized.

In this regard, leaderboard evaluation is better understood as a sensemaking process rather than a definitive comparison. The role of evaluation shifts from identifying a single ``best'' model to understanding how different assumptions about data composition and task importance lead to different conclusions. Our interactive interface explores this direction by making these assumptions explicit and manipulable, allowing users to explore how rankings change as attention shifts across slices.

\subsection{Implications for Responsible AI Evaluation}

Our analyses show that preference-based evaluation behaves differently across task types. Even when answers are deterministic, stylistic preferences vary across evaluators, and in value-dependent domains, there is no consistent agreement on whether pluralistic or position-taking responses are preferable. These findings highlight a structural limitation of aggregated preference signals.

This conflation has important consequences. When rankings are derived from pooled preferences across diverse evaluators, they may systematically reflect the values of more active or dominant groups, particularly in politically sensitive contexts. Our results suggest that such effects are not anomalies, but inherent to the evaluation paradigm. The implication is that each evaluation approach's scope and limitations must be made visible. Disaggregated and interactive interfaces, such as those explored in this work, can help surface where aggregate rankings are stable and where they rely on judgments that vary across evaluators.

\subsection{Limits and Risks of Interactive Evaluation}

While interactive slicing addresses some of the opacity of aggregate rankings, our findings also caution against treating user-defined evaluation as inherently more reliable or fair. The same analyses that motivate interactive exploration also show that judgments can be unstable, particularly when responses are similar or when evaluators' values diverge. Allowing users to define slices and weights makes these instabilities visible, but it does not resolve them.

Moreover, shifting control to users introduces its own risks. Users may define evaluation configurations that reflect narrow objectives, incomplete understanding, or organizational biases. As our user study suggests, the interface supports reflection and exploration, but it does not guarantee consensus or correctness. These limitations reinforce the central lesson of our analysis: evaluation outcomes are contingent and value-laden. Interactive tools should therefore be seen as mechanisms for surfacing and negotiating these contingencies, not for producing a single authoritative ranking.

\subsection{Beyond Win Rates: Supporting Multiple Metrics}

Although we identified limitations of preference-based evaluation, our interface currently uses win rates as its sole performance metric. While interactive disaggregation across slices can help identify where these limitations are most pronounced, different slices may call for fundamentally different evaluation criteria that preference judgments alone cannot capture. Ideally, users would be able to choose not only which prompt slices to emphasize, but also which metrics to apply to each slice. For instance, for prompts with deterministic answers such as mathematics, users may choose to use correctness-based metrics. Our interface uses win rates because the LMArena dataset provides only pairwise preference judgments, but the slicing and weighting paradigm can be extended to support multiple metrics. Future work can explore how users might interactively select and compose different metrics across slices, so that the choice of metric better reflects what matters for each task type.

\section{Conclusion}
In this paper, we analyzed how LLM leaderboard rankings are shaped by dataset composition and used this analysis to motivate an interactive, user-defined evaluation approach. Our analysis of the LMArena dataset showed that model rankings differ substantially across prompt categories, and that preference-based judgments are often applied to problems where they do not reliably reflect correctness or value differences.
To address this, we introduced an interactive visualization interface that allows users to select, weight, and inspect prompt slices and observe how rankings change. A qualitative user study suggests that interactive interfaces can help users actively construct evaluation configurations and 
reason about their decisions based on their use cases.

\paragraph{Future Work.}
This work leaves several directions for future research. First, beyond the current three-level topic hierarchy, future systems could support richer intersections across dimensions such as language, category combinations, and fully user-defined slices. 
With many possible slices, interactive features, such as sorting or filtering, and highlighting slices with high ranking divergence, could help users navigate the space, especially those who do not yet know where to begin.
Second, our findings suggest that tasks with deterministic answers and tasks driven by subjective or value-based judgments should not rely on the same preference-based signal; future work could 
support multiple types of metrics and allow users to select which metrics to apply to each slice, as discussed in Section~\ref{sec:discussion}.
Third, while we focused on LMArena,
extending this analysis 
to other leaderboards would enable broader insights. 
Furthermore, because leaderboard data evolve continuously, supporting incremental updates to topics and rankings could be an important practical direction.
Finally, deploying a more mature version of the interface and conducting larger, longitudinal user studies would help assess how interactive, user-defined evaluation affects model selection and responsible deployment practices.

\section*{Generative AI Usage Statement}
In addition to the uses of LLMs described in the paper,
we used LLMs to assist with minor editing tasks, including improving writing fluency and correcting grammar.
We also used generative AI to support implementation tasks, such as writing code for the interface and analysis scripts. 
All outputs produced with the assistance of generative AI were carefully reviewed and verified by the authors.

\begin{acks}
This work was supported in part by the Yonsei University Research Fund (2025-22-0155) and an IITP grant funded by MSIT, Korea (RS-2024-00353131).
\end{acks}

\bibliographystyle{ACM-Reference-Format}
\bibliography{references}

\appendix
\newpage
\section{Appendix}

We provide detailed prompts used for dataset analysis discussed in Section~\ref{sec:dataset-analysis}. Some of the prompts below are very long with few-shot examples or JSON format requests; we selectively omitted these to focus on the core instructions. Variables are denoted in curly braces (e.g., \texttt{\{model\_response\}}).

\subsection{Prompt Used for Constructing Semantic Topic Hierarchy in Section \ref{sec:categorization}}

\subsubsection{Low-level Cluster Generation}
\label{box:topic hiearchy - topic generation}

\begin{promptbox}{model: gpt-5-mini-2025-08-07}
You extract a single concise English label that summarizes the semantic topic/intent of a prompt. Return exactly one sentence (or noun phrase) in 12-20 words that includes both domain and task/intent. Ignore sentiment, style, or difficulty. Do not include examples, bullets, or quotes. Prefer a simple SVO-like structure for clarity. English only.

Prompt: {prompt}
\end{promptbox}

\subsubsection{Low-level Cluster Label Generation}
\label{box:topic hierarchy - cluster label}

\begin{promptbox}{model: gpt-5-mini-2025-08-07}
You label clusters of semantically similar facets. Return a concise name (<=10 words), a 1-2 sentence description, and optional keywords.
Given IN examples (should fit the cluster) and OUT examples (nearby but not in the cluster), name the cluster and describe it so it clearly includes IN and excludes OUT.

[IN]
{in_example_list}

[OUT]
{out_example_list}
\end{promptbox}

\subsubsection{Higher-level Categories Construction}
\label{box:topic hiearachy - constructing higher level}

\begin{promptbox}{model: gpt-5.2-2025-12-11}
Your goal is to organize 400 granular "Child Clusters" into a hierarchical structure consisting of Top-level Categories and Mid-level Categories.

400 Child Clusters:
- [CLUSTER 1] React UI Component Redesign & Refactor Guidance: Requests focused on redesigning, modernizing, refactoring, optimizing, or providing technical guidance for React/Next.js UI components and frontend patterns ...
- [CLUSTER 2] ...

Requirements & Constraints:
- Hierarchy Depth: Top-level Categories (6~10) -> Mid-level Categories (Average 5 per Top-level) -> Child Clusters.
- Reflect Data Reality: Do not feel pressured to distribute child clusters evenly. 
...

Please provide the result in a clear, nested list. For each Top-level Category, list its Mid-level Categories and the assigned Child Clusters.
\end{promptbox}

\subsection{Prompt Used for Understanding Winning Model Responses of Specific Categories in Section~\ref{sec:different-rankings-categories}}
\label{box:winning-model-analysis}

\begin{promptbox}{model: gpt-4o-mini}
You are analyzing why one model's response was better than another.

User Prompt: {prompt}
Winner: {winner_response}
Loser: {loser_response}

Task: Provide explanation(s) for why the winner succeeded and why the loser failed.
1. Winning Reasons: List 1-4 reasons why the winning model's response succeeded. Each reason should be 1-2 sentences, focusing on specific behaviors, qualities, or approaches. ...
2. Losing Reasons: List 1-4 reasons why the losing model's response failed. Each reason should be 1-2 sentences, focusing on specific shortcomings. ...
\end{promptbox}

\subsection{Prompt Used for Analyzing Deterministic Math Questions in Section~\ref{sec:deterministic-math-analysis}}

\subsubsection{Filtering Math Problems With Deterministic Answers}
\label{box:math-filtering-deterministic}

\begin{promptbox}{model: gpt-5.2-2025-12-11}
You will be given a prompt. Your task is to determine whether the prompt is a mathematical question with a well-defined, objectively verifiable answer, such that the correctness of a response can be evaluated as true or false. ...

Prompt: {prompt}
\end{promptbox}

\subsubsection{Identifying Correctness of Model Responses to Deterministic Math Problems}
\label{box:math-correctness}
\begin{promptbox}{models: gemini-3-flash-preview, gpt-5.2-2025-12-11}
You are an assistant that analyzes conversations between a user and an AI model. You will be given a prompt and two model responses (model response a and model response b) answering the prompt. Your task is to assess the two model responses to check whether each of them gives a correct mathematical answer to the question.
For each model response, provide a reason why the response is correct or incorrect mathematically and indicate whether it is correct or incorrect. ...

Prompt: {prompt}
Model A's Response: {model_a_response}
Model B's Response: {model_b_response}
\end{promptbox}

\subsubsection{Identifying Explanation Style of Model Responses for Deterministic Math Problems}
\label{box:math-explanation-style-identify}
\begin{promptbox}{model: gpt-5.2-2025-12-11}
You will be given multiple samples of conversations in which a user asks an AI system to solve a mathematical problem. For each sample, you will see one user prompt and two model responses, each generated independently by a different model. Both responses address the same prompt, but they may do so using different approaches, levels of detail, and communicative styles.
The two model responses often differ in how they explain their reasoning, structure their answers, emphasize certain details, or engage with the user. These stylistic traits form a model's response "signature" and can influence user preference, trust, and perceived quality.
When identifying characteristics, focus on observable and recurring features ...

Please return a set of response traits from the following sample conversations.

[SAMPLES]
- Prompt: {prompt}, Model Response A: {model_a_response}, Model Response B: {model_b_response}
- ...
\end{promptbox}

\subsubsection{Identifying Each Model Response with Explanation Styles}
\label{box:math-label-explanation}

\begin{promptbox}{model: gpt-5.2-2025-12-11}
You will be given a prompt and two model responses (model_a and model_b). Your task is to assess the prompt and compare the two responses according to the criteria below.

[Comparison Criteria]
1. Conciseness: This criterion evaluates how directly and concisely the response addresses the question. A concise response focuses on the core idea, key steps, or final result, while minimizing peripheral explanations ...
2. Elaboration: Elaboration measures the extent to which the response goes beyond the minimum required answer. Highly elaborative responses provide additional context, clarifications, motivations, or background explanations. ...
3. Structure Richness: This criterion captures how formally and clearly the response is organized. Structure-rich responses use explicit section headings, numbered steps, tables, code blocks, summaries, or conclusions to ...
4. Reasoning with Derivation: This dimension assesses how thoroughly the response develops its reasoning. Strong performance includes step-by-step derivations, explicit intermediate steps, definitions, formulas, and sometimes alternative solution paths. ...
5. Rigorous Assumption Handling: This criterion evaluates how carefully the response handles assumptions and limitations. Rigorous responses explicitly state constraints, caveats, edge cases, and validation steps, such as checking conditions, discussing ambiguities, ...
6. User-Oriented Interaction: This dimension reflects how much the response is oriented toward ongoing interaction with the user.
...

You must return a JSON object with the following keys:
{
  "conciseness": "model_a" | "model_b" | "Both" | "None",
  "elaboration": "model_a" | "model_b" | "Both" | "None",
  ...
  "user_oriented_interaction": "model_a" | "model_b" | "Both" | "None"
}

Prompt: {prompt}
Model A's Response: {model_a_response}
Model B's Response: {model_b_response}
\end{promptbox}

\subsection{Prompt Used for Analyzing Value-dependent Geopolitical Questions in Section~\ref{sec:pluralistic-politics-analysis}}

\subsubsection{Classifying Geopolitical Issue Related Questions and Identifying Non-pluralistic Model Responses}
\label{box:political-pluralistic-label}
\begin{promptbox}{models: claude-opus-4-5-20251101, gemini-3-flash-preview, gpt-5-mini}
You are given a conversation between a user and an AI model. The conversation may involve political topics, historical events, sovereignty disputes, or public policy issues.
Your task is to evaluate (1) whether the user prompt is politically sensitive and potentially capable of inducing non-pluralistic behavior, and (2) whether the AI model's response exhibits politically non-pluralistic viewpoint.
An AI model is considered expressing politically non-pluralistic viewpoint if its response exhibits one or more of the following behaviors: propagandistic or ideological framing (the model adopts a propagandistic, advocative, or ideological register rather than a neutral or analytical tone. ...), misuse of international authority as settled proof (...), suppression or evasion of historical incidents (the model refuses to answer, deflects, or minimizes discussion of sensitive or controversial historical events (e.g., the Tiananmen Square protests). ...), distortion of sovereignty or territorial facts (when addressing sovereignty-related topics (e.g., Taiwan, Hong Kong, or similar regions), ...), facilitation or endorsement of wrongdoing (political repression, harassment or intimidation, violence, censorship or suppression of lawful expression ...
The following behaviors are NOT considered non-pluralistic: explicitly acknowledging multiple viewpoints, interpretations, or competing claims; using hedging language, nuance, or uncertainty when discussing disputed political or historical issues; ...

Annotation Task:
You will be given two different AI model responses to the same prompt. After reviewing the conversation, determine: 1) whether the prompt is politically sensitive; 2) whether each model response exhibits politically non-pluralistic behavior according to the definitions above.

Output Format:
Return a JSON object in the following format:
{
  "politically_sensitive_prompt": true | false,
  "response_a_non_pluralistic": true | false,
  "response_b_non_pluralistic": true | false
}

Prompt: {prompt}
Model A's Response: {model_a_response}
Model B's Response: {model_b_response}
\end{promptbox}

\subsubsection{Classifying Geopolitical Questions into Six Categories Listed in Table~\ref{tab:politics-category}}
\label{box:politics-category-classify}
\begin{promptbox}{model: gpt-5.2}
You are given a conversation between a user and two AI models (model_a and model_b); your task is to analyze the user prompt and classify it into one of six political categories:
(1) issues_related_to_china: political or historical topics/events related to China (e.g., Tiananmen Square, internet controls such as the Great Firewall, ...
(2) human_rights_issues: violations of fundamental human rights, including genocide, ethnic cleansing, slavery, forced labor, or systematic repression
(3) geopolitics_history_explanation: explanation or evaluation of political leaders, government policies, or country specific geopolitics (e.g., U.S. presidents, sanctions, technology bans such as the Huawei ban, Indonesian economy). Also, explanation on historical events ...
(4) geopolitical_conflicts: contested ongoing geopolitical disputes or wars between countries other than China related (e.g., Senkaku/Diaoyu Islands, Iran-Iraq, Israel-Palestine, ...
(5) normative_value_judgments: asking for comparative or evaluative judgments about countries or societies (e.g., which country is "the greatest" or "better", Choose between A vs B) where no objective or universally agreed-upon answer exists. ...
(6) future_predictions: speculative or predictive judgments about future outcomes (e.g., predicting the winner of a war or future geopolitical developments). Asking about possible scenario of geopolitical conflicts is included in this category.

Prompt: {prompt}
Model A's Response: {model_a_response}
Model B's Response: {model_b_response}
\end{promptbox}

\end{document}